# Abusive and Threatening Language Detection in Urdu using Supervised Machine Learning and Feature Combinations

Muhammad Humayoun

*Higher Colleges of Technology, Abu Dhabi, United Arab Emirates*

**Abstract**
This paper presents the system descriptions submitted at the FIRE Shared Task 2021 on Urdu's Abusive and Threatening Language Detection Task. This challenge aims at automatically identifying abusive and threatening tweets written in Urdu. Our submitted results were selected for the third recognition at the competition. This paper reports a non-exhaustive list of experiments that allowed us to reach the submitted results. Moreover, after the result declaration of the competition, we managed to attain even better results than the submitted results. Our models achieved 0.8318 F1 score on Task A (Abusive Language Detection for Urdu Tweets) and 0.4931 F1 score on Task B (Threatening Language Detection for Urdu Tweets). Results show that Support Vector Machines with stopwords removed, lemmatization applied, and features vector created by the combinations of word n-grams for n=1,2,3 produced the best results for Task A. For Task B, Support Vector Machines with stopwords removed, lemmatization not applied, feature vector created from a pre-trained Urdu Word2Vec (on word unigrams and bigrams), and making the dataset balanced using oversampling technique produced the best results. The code is made available for reproducibility[1].

## 1. Introduction

With the exponential growth of social media users, hate speech is becoming a pandemic in which no one is safe. With the luxury of anonymity and virtual space, people tend to say things they may have filtered in a physical setting. Two important types are the use of speech in writing which is (1) abusive and (2) threatening. Rapid growing research is happening to find the methods and techniques that detect such hate speech automatically for English. However, resource-poor languages do not get that much attention. Shared tasks like this are generally considered a good step. Abusive and Threatening Language Detection Task in Urdu[3] is a shared task at CICLing 2021 track at FIRE 2021 [1] [2] co-hosted with ODS SoC 2021[4]. Specifically, two tasks are proposed in this challenge. Task A focuses on detecting Abusive language, whereas, Task B focuses on detecting Threatening language from the tweets written in Urdu. Both can be treated as binary classification problems, and supervised learning models can be employed to train.

This paper presents the system descriptions which were submitted at the competition. Our submitted results were selected for the third recognition with a monetary prize of 10K Rub sponsored through the ODS Summer of Code. This paper reports a non-exhaustive list of experiments that allowed us to reach the submitted results. In addition, we managed to attain better results than the submitted results in the competition. The related research work outside of this competition describing the dataset construction and producing excellent results are reported in [3] [4].

---

[1] https://github.com/humsha/UrduAbusiveandThreatDetectionTasks

[3] https://www.urduthreat2021.cicling.org/home
[4] https://ods.ai/tracks/summer-of-code-2021

There are more than 100 million people worldwide speaking Urdu. It is an Indo-Aryan language, having a modified Perso-Arabic alphabet [5], and written in Nastalique writing style. Urdu draws its advanced vocabulary from Persian and Arabic and day-to-day usage vocabulary from the native languages of South Asia [6]. Urdu lacks capitalization, which makes identifying proper nouns, titles, acronyms, and abbreviations difficult. Similar to Arabic and Persian, vowels are optional and hardly present in the written text. [7] Thus, words are often guessed with the help of context. Urdu is a free word order (Subject Object Verb) language [8].

## 2. Dataset Description

The dataset related to the Abusive task has 2,400 instances in the training set (1,213 instances labeled as non-abusive, and 1,187 instances labeled as abusive). The test set has 1100 instances (non-abusive instances: 537, abusive instances: 563). A cursory analysis of the dataset reveals that many of the abusive instances contain abusive words making the detection straightforward.

The dataset related to the Threatening task has 6K instances in the training set (4,929 instances labeled as neutral and 1,071 instances as threatening). The balance of the classes is 70% (neutral) and (30% threat), which is substantially imbalanced. The test set has 3,950 instances (neutral instances: 3231, threatening instances: 719). In addition, threatening task detection can fundamentally be a difficult problem, mainly due to subtle hints of threat in the threatening instances. Also, there can be unavoidable variation in human judgment regarding an instance in hand, and a difference of opinion may arise. For instance, some of the tweets labeled "threatening" do not seem to be threatening to us as native speakers. Also, a political, religious, and regional affiliation, among others, can add bias to the annotation process. Indeed, this is what makes this task difficult. Figure 1 shows some of such instances and our comments.

**Further Observations:**

A superficial analysis of both datasets reveals extensive use of non-standard script in tweets with a high number of spelling mistakes (which is quite normal with an informal text such as tweets). We also found a high number of non-standard canonical form of words with spelling variations. Proper segmentation of Urdu words is another important problem. We found that a number of words joined together as one token should be separated by space. All these suggest the limitations of applying NLP tools such as removing stopwords, normalization, and applying lemmatization & Word2Vec.

## 3. Preprocessing, Features extraction and Classification Techniques
### 3.1. The Preprocessing

Preprocessing is the first step in NLP, and in particular in text classification. We apply the following preprocessing:

1. **Compulsory Preprocessing.** The following preprocessing are applied to every document.
    a. *Diacritic Removal*. Diacritic marks are not consistently used in Urdu. To ensure the consistency of data, removing all the diacritics is a common practice, though we lose valuable information in this step.
    b. *Text Normalization*. Urdu is written using a modified Perso-Arabic script. Characters that visually look similar have different Unicode resulting in orthographic variations. We normalize all such variations in Normalization Form C (NFC) [9].
2. **Stopword Removal**. The stopword list we used is provided by [10, 11], and it contains nearly 500 words.
3. **Lemmatization**. We used Urdu Morphological Analyzer [5] to convert all the surface forms of a word to its lemma or root. This tool covers approximately 5000 words, capable of handling 140,000 word forms.
4. **Artificial Data Generation**. Synthetic Minority Oversampling Technique (SMOTE) [12] is used to balance the imbalanced dataset in Task B. The minority class in increased by SMOTE over-sample technique.

## 3.2. Feature Extraction

We have used the following word-level and character-level features.
1. **Bag of words model.** A standard bag of words model using a combination of a non-exhaustive list of features produced by character n-grams and word n-grams.
    1.1. **Weighting schemes.** The value of a feature in a bag of word model is calculated using:
        1.1.1. Raw counts
        1.1.2. Binary: words are marked as present (1) or absent (0).
        1.1.3. Frequency: words are scored based on their frequency of occurrence within the document.
        1.1.4. TF-IDF: words are scored based on their frequency, and common words across the documents are penalized.
2. **Selecting K-best features** using *SelectKBest* algorithm available in scikit-learn. We report results with K=1000 and 5000.
3. **Word Embeddings**. Urdu Word Embeddings [13] is a pre-trained Word2Vec implementation for Urdu that we have used in some of our models. It covers is 100K words. The Procedure to form a vector model for the tasks in hand is given in Figure 2.

## 3.3. Classification Techniques

Both traditional machine learning and neural network techniques have been extensively used in the literature for similar tasks [3] [4]. We selected the following machine learning classifiers because of their superior results on the tasks in hand[5]:

1. **Support Vector Machines** with kernels RBF, Sigmoid and Polynomial (with degree 1, 2, 3).
2. **AdaBoost** with default settings.
3. A **Convolution Neural Network** for sentence classification is reported in [14]. Our model is its simplified version. The main difference is that our model does not use pre-trained Embeddings. We define a model with four input channels. In these channels the text is processed with word n-grams (n is 1 to 4) settings. Each channel is defined as:
    3.1. An input layer
    3.2. Embedding layer set to the size of the vocabulary and 100-dimensional real-valued representation.
    3.3. Convolutional layer of 1-dimension with 32 filters and a kernel size set to the number of words to read at once (word n-grams where n=$k$ for channel$_k$ with k=1, 2, 3, 4, i.e. channel$_1$ used unigrams, channel$_2$ used bigrams, and so on).
    3.4. Max Pooling layer to combine the output from the convolutional layer.
    3.5. Flatten layer to reduce the three-dimensional output to two dimensional for concatenation.
4. The output from the four channels are concatenated into a single vector and process by a Dense layer and an output layer.

## 4. Experiments:

For both tasks, train and test sets are given. During the competition, a portion of the test set was made public for the participants to compare the results. The private part of the test set was released after the competition, and final results were announced[6]. The results reported in this paper are from the complete test set (both private and public). The models are produced by training a classifier on training set and the predictions are performed on test set. The experiments are performed on a laptop with processor Intel Core i7 8[th] generation with 8 GB RAM.

---

[5] Naïve Bayes, Logistic Regression, Random Forests, Decision Tree and Multilayer Perceptron were excluded due to space limitations and somewhat mediocre results.
[6] https://www.urduthreat2021.cicling.org/

## 4.1. Abusive language detection task (Task A)

**Experiment A1:** In this experiment, we produce a bag of word feature vector with an exhaustive list of combinations between removal of stopwords (yes/no), applying lemmatization (yes/no), word n-grams (n=1, 2, 3) and modes (freq, count, binary, tfidf). Character level n-grams are not included in this experiment in order to keep the vector size reasonable for the laptop. The top results greater than 80% F1 sore are shown in Table 1. The results reported in section are the best ones among all the experiments for Task 1.

**Experiment A2:** In addition to the feature combinations in Experiment A1, we include character n-grams of n=1, 2, 3. Then, K-best features for K=1000 and 5000 are selected. Finally, the traditional classifiers are applied for training and prediction. The top results with F1 greater 82% are shown in Table 2.

**Experiment A3:** The feature vector for each comment is produced by a pre-trained Urdu Word2Vec [13] using the procedure given in Figure 2. The feature vector is produced for word n-grams n=1,2. Then, the traditional classifiers are applied for training and prediction. The results do not improve further from 79% as shown in Table 3. As experiment A1 demonstrates that removing stopwords and applying lemmatization produces among the highest results, we only report these preprocessing settings in Table 3.

**Experiment A4:** A model for Convolution Neural Network (CNN) is discussed in Section 3.3. In this experiment, we apply this model on the dataset. As preprocessing, stopwords are removed, and lemmatization is applied. As discussed in Section 3.3, word n-grams for n=1,2,3,4 are applied in 4 channels. The vocabulary size is ~25K, and the vector size is 300. As the model is stochastic, an average of 5 runs of training and prediction is reported in Table 4.

## 4.2. Threatening language detection task (Task B)

**Experiment Task B1:** As mentioned in Section 2, the dataset for Urdu's threatening language detection task is substantially imbalanced. Most of the experiments we performed for Task A did not produce reasonable results when applied to for task B. However, when an additional step of balancing the dataset is added, we get somewhat good results as described below:

The feature vector for each comment is produced using the procedure given in Figure 2. It means that we employed the pre-trained Urdu Word2Vec to generate the feature vector for each tweet in the dataset (the process is similar to experiment A3). The feature vector is produced for word n-grams n=1,2. The dataset is balanced by over sampling using SMOTE [12]. Finally, the classifiers are applied for training and prediction. Note that a non-exhaustive list of combinations of preprocessing settings (stopwords and lemmatization) is applied, but only top results with F1 greater than 47% are reported in Table 5. Character n-grams are not used as we cannot get vectors for characters from Urdu Word2Vec.

**Experiment Task B2:** In this experiment, we produce a bag of word feature vector. The stopwords are removed, and lemmatization is applied; combinations of word n-grams (n=1, 2) and mode ('freq') is used. The dataset's bigger size (unigrams and bigrams together for the last two experiments in Table 6) exhausted the 8GB RAM. Therefore, we reduced the vocabulary size for these two experiments by only considering the tokens that occur at least 4 times in the dataset. The dataset is balanced by oversampling using SMOTE. By looking at the mediocre results we do not see any point to produce an exhaustive list of preprocessing combinations. The results are reported in Table 6.

**Experiment Task B3:** Another way of reducing the size of feature vector is to select top K features. For K=2000, the results are shown in Table 7. Stopwords are removed and lemmatization is applied. Combinations of word and character n-grams (n=1, 2) and modes ('freq', 'tfidf') are used. It is noted that the score started to decrease when character n-grams n=2 is added to the word bigrams n=1,2 in the last two experiments of Table 7. Therefore, we did not complete an exhaustive list of settings or go for n=3 for word and character n-grams.

**Experiment Task B4:** In this experiment, we apply the Convolution Neural Network (CNN) model discussed in Section 3.3 on the dataset. As preprocessing, stopwords are removed, and lemmatization is applied. The dataset is balanced by oversampling using SMOTE. Finally, CNN is applied. As discussed in Section 3.3, word n-grams for n=1,2,3,4 are used in 4 channels. The vocabulary size is ~11K, and the vector size is 300. As the model is stochastic, an average of 5 runs of training and prediction is reported in Table 8. However, the results are not good.

## 5. Discussion on Results

For the Abusive language task (Task A), the traditional models (Bag of words with SVM) performed best in various settings. It is understandable as the dataset is small and CNN probably needs more data to outperform the traditional models. The use of pre-trained Urdu Word2Vec in experiment A3 also underperforms. It is probably because of the limited vocabulary of the pre-trained Word2Vec and the non-standard orthographic variations found in the tweets. Experiment A1 also reveals that increasing the feature set does not necessarily increase the results (though the results remain comparable but expensive computationally). It turns out that removing stopwords and applying lemmatization is a reliable preprocessing. Combining word level unigrams, bigrams and trigrams produce results among the best results. Combining more n-grams increases the sentence vector size enormously and should be avoided when computational power is scarce. Though, unigrams with bigram performed the highest scoring results for task A (See Table 1). In terms of weighting scheme, "freq" turns out to be reasonably reliable. Similarly, SVM with polynomial degree 1, sigmoid, and RBF turns out to be reasonably reliable among various settings.

As shown in experiment A2 (Table 2), combining word level and character level unigrams, bigrams, and trigrams first and then selecting top 1K to 5K features is also a suitable method producing 80%+ results. In this case, it seems that keeping the stopword and not applying lemmatization still produces the best results in most cases. It seems that keeping the top features from a huge feature set of word and character level N-grams with n=1,2,3 manages to discover lexical patterns.

The dataset for the threatening language detection task (Task B) is highly imbalanced. Most of the experiments we performed for Task A did not produce good results on Task B. One of the main points worth noting is the need to make the dataset balanced. It seems that oversampling has increased the scores, but the best score reported is still below 50% for F1 score. The reason might be the subtle nature of the task at hand, as discussed in Section 2.

The highest F1 score is found in experiment B1 when the feature vector is produced by a pre-trained Word2Vec using (word n-grams for n=1,2). We think that oversampling by SMOTE managed to create the feature vectors for new instances in such a way that a somewhat separation of labels was possible. Results in all other experiments are even worse. Especially the results for CNN with or without oversampling are very low.

## 6. Conclusion

In this paper, we have described the systems that produced the reported results for the competition. Moreover, after the result declaration of the competition, we managed to attain even better results than the submitted results. Our analysis of the results shows that until large enough datasets are made available; we can rely on (1) the traditional Bag of word models to create features and (2) the conventional classifiers such as SVM. Effective training of Neural Network-based methods can catch up quickly only if a large training set is readily available. We found Task B quite challenging due to the inherent shortcomings of an imbalanced dataset and the difficulty of the task at hand. The shortcoming of the oversampling technique that we employed is witnessed. Generally, oversampling is a challenging task for NLP datasets. We learnt that having a balanced and large enough dataset is an important prerequisite for better results though such datasets might be hard to produce. As a future work, we shall explore the recent transfer-learning techniques such as BERT fine-tuning, though getting a large enough Urdu BERT model might be a challenge.

## Acknowledgment

We thank the anonymous reviewers for their insightful comments and suggestions.

# Appendix

| ID | Tweet | Label |
|---|---|---|
| 1594 | ان بیغیرتوں کا یہی علاج ہے، جتنی زیادہ دیر رہیں گے اتنا زیادہ گند ماریں گے۔ | 1 |
|  | **English**: This is the cure for these shameless people. The longer they stay, the more they will create mess.<br>**Comment**: Certainly Abusive but only a subtle hint of threat. (It is more like a threatening prediction). Author agrees with the label. |  |
| 1533 | قومی غیرت ایک ایسی چیز ہے جو ملک و قوم کی دنیا بھر میں پہچان بناتی ہے، آج ہندوستان نے کشمیر پر تمام دنیاوی قانون جو۔۔۔ | 1 |
|  | **English**: National pride is something that makes the reputation of a nation worldwide. Today, India has imposed all worldly laws on Kashmir which ...<br>**Comment**: Not threatening; not abusive. Author disagrees with the label |  |
| 413 | تاریخ ضرور لکھے گی غدار کون تھا اچھے بھلے ملک کا بیڑا غرق کیسے اور کس نے کیا۔ | 1 |
|  | **English**: History will surely write who was a traitor, how and who destroyed the prosperous country<br>**Comment**: Not threatening; not abusive. Author disagrees with the label |  |
| 2150 | میں ایک مسلم لیگی کی حیثیت سے اس کی مذمت کرتا ہوں۔ | 1 |
|  | **English**: I condemn it as a member of Muslim League<br>**Comment**: Not threatening; not abusive. Author disagrees with the label |  |
| 2379 | ناکوں چنے چبوائے اور آخر میں ذلیل کروا کر افغانستان سے نکلوایا۔ | 1 |
|  | **English**: They were given a tough time and finally expelled from Afghanistan with humiliation.<br>**Comment**: Not threatening; not abusive. Author disagrees with the label |  |

**Figure 1:** Some Sample tweets from Threatening language dataset

```
x_doc= [token₁, token₂, …, tokenₙ] where tokenᵢ is a space separated chunk of string
x_docs= [x_doc₁, x_doc₂, …, x_docₙ] where x_docᵢ is a complete comment
doc_vectors = [doc_vector₁, doc_vector₂, …, doc_vectorₙ]
                              where doc_vectorᵢ is document vector of x_docᵢ
for each doc in x_docs:
    temp = a 300-dimension vector of numbers
    for each word in doc:
          word_vec = a 300-dimension vector for word from embedding
          add word_vec in temp
    end for
    doc_vector = (aggregate n vectors from temp into one vector
                              of size 300 using a column-wise average)
    add doc_vector in doc_vectors
end for
return doc_vectors
```

**Figure 2**: Procedure to form a vector model from Word2Vec

**Table 1**
Abusive language detection Task A1: Bag of words with traditional classifiers – Top results greater than 80% F1 score

| STW Removed | Lemmatized | Word ngrams | Mode | Model | Vocab size | Vector size | F1 |
|---|---|---|---|---|---|---|---|
| Yes | Yes | 1,2 | freq | SVM-POLY-1 | 24949 | 18963 | 0.8318 |
| Yes | Yes | 1 | freq | SVM-POLY-1 | 5695 | 4657 | 0.8312 |
| Yes | Yes | 1,2,3 | freq | SVM-POLY-1 | 45014 | 33596 | 0.8245 |
| Yes | Yes | 1 | freq | SVM-SIGMOID | 5695 | 4657 | 0.8231 |
| Yes | Yes | 1,2 | freq | SVM-SIGMOID | 24949 | 18963 | 0.8218 |
| Yes | Yes | 1,2,3 | freq | SVM-SIGMOID | 45014 | 33596 | 0.8209 |
| Yes | Yes | 1,2 | freq | SVM-RBF | 24949 | 18963 | 0.819 |
| Yes | Yes | 1,2 | count | SVM-SIGMOID | 24949 | 18963 | 0.8174 |
| Yes | Yes | 1,2 | binary | SVM-SIGMOID | 24949 | 18963 | 0.8165 |
| Yes | Yes | 1,2 | count | SVM-RFB | 24949 | 18963 | 0.8153 |
| Yes | Yes | 1,2 | count | SVM-POLY-1 | 24949 | 18963 | 0.815 |
| **Yes** | **No** | **1** | **freq** | **SVM-SIGMOID** | **6711** | **5466** | **0.8144**[7] |
| Yes | Yes | 1,2,3 | binary | SVM-RFB | 45014 | 33596 | 0.8127 |
| Yes | Yes | 1,2 | binary | SVM-RFB | 24949 | 18963 | 0.8126 |
| Yes | Yes | 1,2 | tfidf | SVM-POLY-1 | 24949 | 18963 | 0.8115 |
| Yes | Yes | 1,2,3 | tfidf | SVM-POLY-1 | 45014 | 33596 | 0.8104 |
| No | No | 1 | binary | SVM-SIGMOID | 6955 | 5707 | 0.8102 |
| Yes | Yes | 1,2,3 | count | SVM-SIGMOID | 45014 | 33596 | 0.8101 |
| Yes | Yes | 1,2,3 | binary | SVM-SIGMOID | 45014 | 33596 | 0.8098 |
| No | No | 1 | binary | SVM-POLY-1 | 6955 | 5707 | 0.8098 |
| Yes | Yes | 1,2,3 | freq | SVM-RBF | 45014 | 33596 | 0.8095 |
| No | No | 1 | tfidf | SVM-SIGMOID | 6955 | 5707 | 0.809 |
| Yes | Yes | 1,2 | binary | SVM-POLY-1 | 24949 | 18963 | 0.8089 |
| No | No | 1 | count | SVM-SIGMOID | 6955 | 5707 | 0.8084 |
| Yes | Yes | 1,2,3 | count | SVM-RBF | 45014 | 33596 | 0.8082 |
| No | No | 1 | count | SVM-POLY-1 | 6955 | 5707 | 0.8081 |
| Yes | Yes | 1,2 | tfidf | SVM-SIGMOID | 24949 | 18963 | 0.8068 |
| No | No | 1 | tfidf | SVM-POLY-1 | 6955 | 5707 | 0.8065 |
| Yes | Yes | 1,2 | count | AdaBoost | 24949 | 18963 | 0.8063 |
| Yes | Yes | 1,2 | tfidf | AdaBoost | 24949 | 18963 | 0.8063 |
| Yes | Yes | 1,2,3 | count | SVM-POLY-1 | 45014 | 33596 | 0.8063 |
| Yes | Yes | 1,2,3 | count | AdaBoost | 45014 | 33596 | 0.8057 |
| Yes | Yes | 1,2,3 | tfidf | AdaBoost | 45014 | 33596 | 0.8054 |
| No | No | 1 | freq | SVM-SIGMOID | 6955 | 5707 | 0.8054 |
| Yes | Yes | 1,2,3 | tfidf | SVM-SIGMOID | 45014 | 33596 | 0.8048 |
| No | No | 1 | freq | SVM-POLY-1 | 6955 | 5707 | 0.8028 |

---

[7] This result is reported to the competition for Task A. There is a negligible variation because of slight changes made in the stopword list.

**Table 2**

Abusive language detection Task A2: Bag of words with traditional classifiers – Top results with F1 greater than 82% (Both word and character level n-grams included for n=1,2,3)

| STW Removed | Lemmatized | Mode | Model | Vocab size | Vector size | Top K | F1 |
|---|---|---|---|---|---|---|---|
| No | No | count | SVM-POLY-1 | 135747 | 104243 | 5000 | 0.8282 |
| No | No | freq | SVM-POLY-1 | 135747 | 104243 | 1000 | 0.8277 |
| Yes | Yes | binary | SVM-RBF | 96923 | 76110 | 1000 | 0.8254 |
| No | No | count | SVM-POLY-1 | 135747 | 104243 | 1000 | 0.8251 |
| No | No | binary | SVM-RBF | 135747 | 104243 | 1000 | 0.8236 |
| Yes | Yes | freq | SVM-POLY-1 | 96923 | 76110 | 1000 | 0.8232 |
| No | No | freq | SVM-POLY-1 | 135747 | 104243 | 5000 | 0.8231 |
| No | No | binary | SVM-POLY-1 | 135747 | 104243 | 1000 | 0.8229 |
| No | No | tfidf | SVM-RBF | 135747 | 104243 | 5000 | 0.8226 |
| No | No | binary | SVM-POLY-1 | 135747 | 104243 | 5000 | 0.822 |
| No | No | tfidf | SVM-RBF | 135747 | 104243 | 1000 | 0.8206 |
| Yes | Yes | tfidf | SVM-POLY-1 | 96923 | 76110 | 1000 | 0.8202 |
| Yes | Yes | binary | SVM-POLY-1 | 96923 | 76110 | 1000 | 0.8198 |
| No | No | freq | SVM-RBF | 135747 | 104243 | 5000 | 0.8196 |
| No | No | count | SVM-RBF | 135747 | 104243 | 1000 | 0.8188 |
| Yes | Yes | tfidf | SVM-RBF | 96923 | 76110 | 1000 | 0.8185 |
| No | No | binary | SVM-RBF | 135747 | 104243 | 5000 | 0.8181 |
| Yes | Yes | freq | SVM-RBF | 96923 | 76110 | 1000 | 0.8171 |
| No | No | tfidf | SVM-POLY-1 | 135747 | 104243 | 1000 | 0.8153 |
| Yes | Yes | binary | AdaBoost | 96923 | 76110 | 1000 | 0.8146 |
| No | No | tfidf | SVM-POLY-1 | 135747 | 104243 | 5000 | 0.8136 |
| No | No | count | SVM-RBF | 135747 | 104243 | 5000 | 0.8114 |
| No | No | freq | SVM-RBF | 135747 | 104243 | 1000 | 0.8113 |
| Yes | Yes | tfidf | AdaBoost | 96923 | 76110 | 1000 | 0.81 |
| No | No | count | AdaBoost | 135747 | 104243 | 5000 | 0.8045 |
| Yes | Yes | freq | AdaBoost | 96923 | 76110 | 1000 | 0.8038 |

**Table 3**

Abusive language detection Task A3: Feature vector by Word2Vec using (word n-grams for n=1,2) with traditional classifiers

| Model | Vocab size | Vector size | F1 |
|---|---|---|---|
| SVM-POLY-1 | 24949 | 300 | 0.7805 |
| SVM-POLY-2 | 24949 | 300 | 0.7912 |
| SVM-POLY-3 | 24949 | 300 | 0.7855 |
| SVM-RBF | 24949 | 300 | 0.7916 |
| SVM-SIGMOID | 24949 | 300 | 0.7579 |

**Table 4**

Abusive language detection Task A4: CNN with 4 channels

|  | F1 |
|---|---|
| Average | 0.797 |
| Maximum | 0.812 |
| Minimum | 0.7804 |
| Std. dev. | 0.89 |

**Table 5**

Threatening language detection Task B1: Feature vector by Word2Vec using (word n-grams for n=1,2) with traditional classifiers. Top results with F1 greater than 47%.

| STW Removed | Lemmatized | Word ngrams | Model | Vocab size | Vector size | F1 |
|---|---|---|---|---|---|---|
| Yes | No | 1,2 | SVM-POLY-3 | 79885 | 300 | 0.4931 |
| Yes | No | 1,2 | SVM-POLY-2 | 79885 | 300 | 0.4921 |
| Yes | No | 1,2 | SVM-RBF | 79885 | 300 | 0.4883 |
| **Yes** | **No** | **1** | **SVM-POLY-3** | **15009** | **300** | **0.487[8]** |
| No | No | 1,2 | SVM-POLY-3 | 102586 | 300 | 0.4866 |
| Yes | Yes | 1,2,3 | SVM-POLY-3 | 145299 | 300 | 0.485 |
| No | No | 1,2 | SVM-RBF | 102586 | 300 | 0.484 |
| No | No | 1,2 | SVM-POLY-2 | 102586 | 300 | 0.481 |
| Yes | No | 1 | SVM-RBF | 15009 | 300 | 0.4778 |
| Yes | Yes | 1,2,3 | SVM-POLY-2 | 145299 | 300 | 0.4766 |
| Yes | Yes | 1,2 | SVM-POLY-2 | 75244 | 300 | 0.475 |
| Yes | Yes | 1,2 | SVM-POLY-3 | 75244 | 300 | 0.4702 |

**Table 6**

Threatening language detection Task B2: Feature vector by bag of word using (word n-grams for n=1,2) with traditional classifiers.

| STW Removed | Lemmatized | Word ngrams | Mode | Model | Vocab size | Vector size | F1 |
|---|---|---|---|---|---|---|---|
| Yes | Yes | 1 | freq | SVM-POLY-1 | 12803 | 9494 | 0.4749 |
| Yes | Yes | 1 | freq | SVM-SIGMOID | 12803 | 9494 | 0.4503 |
| Yes | Yes | 1 | freq | SVM-POLY-2 | 12803 | 9494 | 0.4231 |
| Yes | Yes | 1 | freq | SVM-RBF | 12803 | 9494 | 0.3886 |
| Yes | Yes | 1 | freq | SVM-POLY-3 | 12803 | 9494 | 0.3774 |
| Yes | Yes | 1,2 | freq | SVM-SIGMOID | 55179 | 36036 | 0.3665 |
| Yes | Yes | 1,2 | freq | SVM-RBF | 55179 | 36036 | 0.1814 |

**Table 7**

Threatening language detection Task B2: Feature vector by bag of word using (word n-grams for n=1,2 and character n-grams for n=2) with traditional classifiers. Top K features selected and used.

| Word ngrams | Char ngrams | Mode | Model | Vocab size | Vector size | Top K | F1 |
|---|---|---|---|---|---|---|---|
| 1,2 | Not used | tfidf | SVM-SIGMOID | 55179 | 36036 | 2000 | 43.49 |
| 1,2 | Not used | freq | AdaBoost | 55179 | 36036 | 2000 | 43.22 |
| 1,2 | Not used | freq | SVM-POLY-1 | 55179 | 36036 | 2000 | 43.09 |
| 1,2 | Not used | tfidf | AdaBoost | 55179 | 36036 | 2000 | 42.71 |
| 1,2 | 2 | tfidf | SVM-POLY-1 | 58043 | 38277 | 2000 | 42.6 |
| 1,2 | 2 | tfidf | SVM-RBF | 58043 | 38277 | 2000 | 42.42 |

**Table 8**

Threatening language detection Task B4: CNN with 4 channels

|  | Without SMOTE F1 | With SMOTE F1 |
|---|---|---|
| Average | 34.21 | 34.52 |
| Maximum | 35.47 | 35.28 |
| Minimum | 32.26 | 33.7 |
| Standard deviation | 1.23 | 0.61 |

---

[8] This result is reported to the competition for Task B. There is a negligible variation because of slight changes made in the stopword list.